%% file: main.tex
\documentclass[10pt,twocolumn,letterpaper]{article}

\usepackage{iccv}
\usepackage{times}
\usepackage{epsfig}
\usepackage{graphicx}
\usepackage{amsmath}
\usepackage{amssymb}
\usepackage{amsthm}

\usepackage[pagebackref=true,breaklinks=true,letterpaper=true,colorlinks,bookmarks=false]{hyperref}
\usepackage{subfigure}
\newcommand{\tabincell}[2]{\begin{tabular}{@{}#1@{}}#2\end{tabular}}

\iccvfinalcopy 

\def\iccvPaperID{5797} 

\ificcvfinal\fi
\begin{document}

\title{Improved Techniques for Training Adaptive Deep Networks}

\author{
Hao Li$^1$$^*$\hspace{1.0cm}Hong Zhang$^2$\thanks{First two authors contributed equally}\hspace{1.0cm}Xiaojuan Qi$^3$\hspace{1.0cm}Ruigang Yang$^{2}$\hspace{1.0cm}Gao Huang$^{1}$\thanks{Corresponding author}
\vspace{0.3cm}\\  
$^1$Tsinghua University \quad $^2$Baidu Inc. \quad $^3$University of Oxford 
\\
{\tt\small \{lihaothu,fykalviny,qxj0125\}@gmail.com} \\
\vspace{-0.02in}
{\tt\small yangruigang@baidu.com \quad gaohuang@tsinghua.edu.cn}
}

\maketitle

\input{abs.tex}

\input{intro.tex}

\input{related.tex}
\input{method.tex}

\input{exp.tex}

\input{conclusion.tex}
\paragraph{Acknowledgements.} Gao Huang is supported in part by Beijing Academy of Artificial Intelligence under grant BAAI2019QN0106. Hao Li is supported in part by Tsinghua University Initiative Scientific Research Program and Tsinghua Academic Fund for Undergraduate Overseas Studies. We thank Danlu Chen for helpful discussions.

{\small
\bibliographystyle{ieee_fullname}
\bibliography{citations}
}

\end{document}

%% file: abs.tex

\begin{abstract}
 Adaptive inference is a promising technique to improve the computational efficiency of deep models at test time. In contrast to static models which use the same computation graph for all instances, adaptive networks can dynamically adjust their structure conditioned on each input. While existing research on adaptive inference mainly focuses on designing more advanced architectures, this paper  investigates how to train such networks more effectively. Specifically, we consider a typical adaptive deep network with multiple intermediate classifiers. We present three techniques to improve its training efficacy from two aspects: 1) a Gradient Equilibrium algorithm to resolve the conflict of learning of different classifiers; 2) an Inline Subnetwork Collaboration approach and a One-for-all Knowledge Distillation algorithm to enhance the collaboration among classifiers. On multiple datasets (CIFAR-10, CIFAR-100 and ImageNet), we show that the proposed approach consistently leads to further improved efficiency on top of state-of-the-art adaptive deep networks.

\end{abstract}

%% file: intro.tex

\section{Introduction}

Convolutional neural networks~(CNNs) have gained remarkable success on a variety of visual recognition tasks \cite{alexnet,resnet,fcn,he2017mask}. Modern CNNs such as GoogleNet~\cite{szegedy2015going}, ResNet~\cite{resnet} and DenseNet~\cite{huang2017densely} are endowed with unprecedented network depth to achieve state-of-the-art accuracy. However, very deep models usually come along with high computational cost, which prevents them from performing real-time inference on resource-constrained platforms like smart phones, wearable devices and drones.

Extensive efforts have been made to improve the inference efficiency of deep CNNs in recent years. Popular approaches include efficient architecture design \cite{sandler2018mobilenetv2,ma2018shufflenet,huang2018condensenet,zoph2018learning}, network pruning \cite{han2015deep, li2016pruning, liu2017learning}, weight quantization \cite{chen2016compressing,han2015deep,hubara2016binarized} and adaptive inference \cite{graves2016adaptive,huang2018multi,bolukbasi2017adaptive,veit17conditionalsimilarity,figurnov2016spatially,teja2018hydranets}. Among them, adaptive inference is gaining increasing attention recently, due to its remarkable advantages. First, it is compatible with almost all the other approaches, i.e., adaptive inference can be performed on highly optimized architectures like MobileNets \cite{sandler2018mobilenetv2} and ShuffleNets \cite{ma2018shufflenet}, and can also benefit from model pruning and weight quantization. Second, by conditioning the computation of a deep model on its inputs, adaptive inference can save a considerable amount of computational cost on ``easy'' samples and/or less important regions, drastically reducing the average inference time. Third, adaptive inference algorithms usually have a set of tunable parameters that dynamically control the accuracy-speed tradeoff. This is a valuable property in many scenarios, where the computational budget may change over time or vary across different devices. In contrast to static models which have a fixed computational cost, adaptive models are able to trade accuracy for speed or vise verse on-the-fly, to meet the dynamically changing demand.

Existing works on adaptive inference in the context of deep learning mainly focus on designing more specialized network architectures \cite{huang2018multi,teerapittayanon2016branchynet,teja2018hydranets} or better inference algorithms \cite{bolukbasi2017adaptive}. In comparison, less effort has been made to improve the training process. But in fact, adaptive CNNs usually have quite different architectures as conventional deep models, and training strategies optimized for the later may not be optimal for adaptive models. In this paper, we consider a representative type of adaptive models that have multiple intermediate classifiers at different depths of the network. With this architecture, adaptive computation can be performed by early exiting ``easy'' samples to speed up the inference. 
The multi-scale dense network (MSDNet) proposed in \cite{huang2018multi} represents the state-of-the-art of this type of models. We aim to improve the training efficacy of such multi-exits networks from the following two perspectives.

First, we need to resolve the conflict while jointly optimizing all the classifiers. It has been observed in \cite{huang2018multi} that the individual classifiers in the network tend to negatively affect the learning of one other, and \cite{song2018collaborative} discussed that the backbone network may not converge well due to the accumulataion of gradients from several classifiers in a multi-head neural network. By introducing dense connections, MSDNet has addressed the problem that early classifiers interfere with later ones. However, deep classifiers may also negatively affect earlier classifiers. To this end, we present a \emph{Gradient Equilibrium (GE)} technique that re-scale the magnitude of gradients along its backward propagation path. This allows the gradient to have a constant scale across the network, which helps to reduce gradient variance and stabilize the training procedure.

Second, we aim to encourage collaboration among different classifiers. This is achieved by introducing two modules, named \emph{Inline Subnetwork Collaboration (ISC)} and \emph{One-for-all Knowledge Distillation (OFA)}, respectively. The motivation for ISC is that later exits may benefit from the prediction of early exits. Therefore, we use the prediction logits from previous stage as a prior to facilitate the learning of current and subsequent classifiers.
The OFA follows from the intuition that the last exit always yield the highest accuracy among all the classifiers, and thus it could serve as a teacher model whose knowledge could be distilled into earlier exits.

We conduct extensive experiments on three image-classification datasets~(CIFAR-10, CIFAR-100, and ImageNet). The experiments demonstrate that the proposed techniques consistently improve the efficiency of state-of-the-art adaptive deep networks.

%% file: related.tex
\section{Related Work}

\paragraph{Computationally Efficient Deep Networks.}
 Approaches to computationally efficient deep networks can be summarized as follows.
 One stream focuses on designing efficient network architectures~\cite{sandler2018mobilenetv2,ma2018shufflenet,huang2018condensenet,zoph2018learning,howard2017mobilenets}, including depth-wise separable convolution~\cite{sandler2018mobilenetv2}, point-wise group convolution with channel shuffling~\cite{zhang2018shufflenet}, and learned group convolution~\cite{huang2018condensenet}, to name a few. 
The other line of research explores methods to prune~\cite{han2015deep,li2016pruning,he2018amc} or quantize ~\cite{chen2016compressing,han2015deep,hubara2016binarized} neural network weights.
These strategies are effective when neural networks have a substantial amount of redundant weights, which can be safely removed or quantized without sacrificing accuracy.

\paragraph{Adaptive Inference.}

Recently, a new emerging direction which employs adaptive learning for efficient inference has drawn increasing research attention, with representative works proposed in~\cite{huang2018multi,bolukbasi2017adaptive,lin2017runtime,veit17conditionalsimilarity,kong2018pixel,figurnov2016spatially,li2017dynamic,ying2017depth,wang2018skipnet,mcintosh2018recurrent,wu2018blockdrop,kang2017incorporating}.
Adaptive inference aims at achieving efficient resource allocation during the inference stage without sacrificing accuracy, by strategically save computation on ``easy'' samples.
Compared with other directions to improve network efficiency, adaptive inference gains advantages due to its compatibility, flexibility and high performance. 

Most prior works are dedicated to learning adaptive network topology selection policies. 
Bolukbasi \etal \cite{bolukbasi2017adaptive} adopted an ensemble model with multiple deep networks of varying size, and proposed to learn an adaptive decision function to determine in which stage the example should exit.
Huang \etal \cite{huang2018multi} designed a novel multi-scale convolutional network with multiple intermediate classifiers with various computational budgets, which can be adaptively selected during the inference stage.
On top of ResNet~\cite{resnet} architecture, 
Veit \etal \cite{veit17conditionalsimilarity} and Wang \etal~\cite{wang2018skipnet} designed  gating functions to dynamically choose layers for efficient inference. Figurnov \etal \cite{figurnov2016spatially} further made the gating policy adaptive to spatial locations.
To further enable adaptive inference in pixel-labeling tasks, pixel-wise attentional gating (PAG)~\cite{kong2018pixel} was introduced in~\cite{kong2018pixel} for adaptively selecting a subset of spatial locations to process, and an RNN architecture was proposed in~\cite{mcintosh2018recurrent} for dynamically determining the number of RNN steps according to allowed time budget.
Adaptive inference was also studied in accelerating visual tracking systems in~\cite{ying2017depth}.

Almost all the prior works focus on designing network architectures or algorithms for adaptive inference. In this work, we made an orthogonal effort by exploring effective strategies to facilitate the training of deep networks for adaptive inference.
Our method is model-agnostic and can be applied to various adaptive inference architectures with multiple intermediate classifiers including~\cite{huang2018multi,bolukbasi2017adaptive,mcintosh2018recurrent}

\paragraph{Knowledge Distillation.}
Our work is also related to knowledge distillation strategies explored in~\cite{hinton2015distilling,lan2018knowledge,bucilu2006model,ba2014deep}, 
in which outputs from teacher networks are utilized to supervise the training of student networks. Different from previous trials in training separate teacher models in advance, Lan~\etal~\cite{lan2018knowledge} proposed a one-stage online distillation framework by utilizing multi-branch network ensemble to enhance the target network.
Our \emph{One-for-all Knowledge Distillation} strategy also shares a spirit similar to these knowledge distillation strategies by acquiring knowledge from larger models with higher accuracy. 
However, in contrast to previous work deploying knowledge distillation to obtain better student models, we show such strategies can promote collaborations between multi-scale classifiers inside one single network, and improve the efficacy of adaptive inference.

    
    

%% file: method.tex

\begin{figure*}[bpt]
   \begin{center}
     \includegraphics[width=0.95\linewidth]{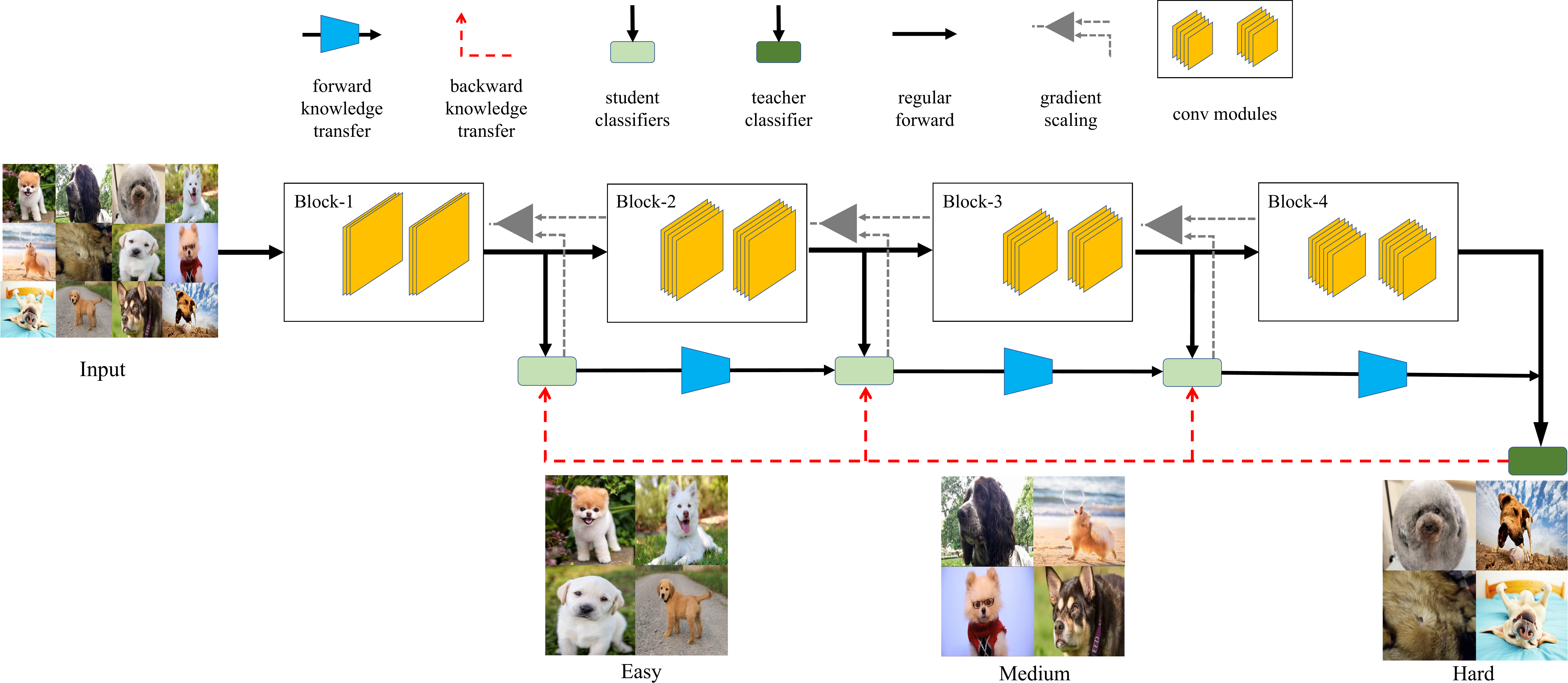}
   \end{center}
   \vspace{-2ex}
      \caption{Overview of the proposed training strategies for adaptive inference on a deep convolutional network. \textbf{Gradient Equilibrium (GE)} is applied to resolve the gradients conflict among different exits. \textbf{Inline Subnetwork Collaboration (ISC)} and \textbf{One-for-all Knowledge Distillation (OFA)} are proposed to enhance the collaboration of different classifiers.}
      \vspace{-2ex}
   \label{fig:overview}
\end{figure*}

\section{Method}

In this section, we first set up the adaptive inference model with multiple exits (classifiers). Then we discuss our proposed techniques for improving the training in detail. Roughly, the first technique, \textit{Gradient Equilibrium (GE)}, is introduced to resolve the gradients conflict of different classifiers; the second and third techniques, \textit{Inline Subnetwork Collaboration (ISC)} and \textit{One-for-all Knowledge Distillation (OFA)}, are proposed to enhance the collaboration among classifiers. Figure \ref{fig:overview} gives an overview of the proposed approach.

\paragraph{Adaptive inference model.} We set up the adaptive inference model as a network that is composed of $k$ classifiers. The model can be viewed as a conventional CNN with $k-1$ intermediate classifiers attached at varying depths of the network. Each classifier is also referred to as an \emph{exit}. 
The model can generate a set consisting of $k$ predictions, one from each of the exits, i.e., 
\[
[y_1, ..., y_k] \!=\! f(x ; \theta) = [f_1(x ; \theta_1), \ldots, f_k(x ; \theta_k)],
\]
where $x$ is the input image, and $f_i$ and $\theta_i$ ($i=1,\dots, k$) represent the transformation learned by the $i$-th classifier and its corresponding parameters, respectively. Note that $\theta_i$'s have shared parameters here.


At test time, the inference is performed dynamically conditioned on each input. Formally, the prediction for a test sample $x$ is given by $\hat y = f_{I(x)}(x, \theta_{I(x)})$, where $I(x) \in \{1,\ldots,k\}$ is a function of $x$, which is usually obtained by a certain decision function. In our experiment, we  simply follow \cite{huang2018multi} to use a confidence-based approach to compute $I(x)$. 

\subsection{Gradient Equilibrium}
Adaptive inference can be considered as a sequential prediction process by a set of subnetworks. A straightforward way to train an adaptive network is to train the subnetworks sequentially. However, this method is far from optimal due to the conflict between two optimization goals: to learn discriminative features for the current classifier, and to maintain necessary information for generating high-quality features for later classifiers~\cite{huang2018multi}. A more effective training strategy is to jointly optimize all the subnetworks. For example, the MSDNet~\cite{huang2018multi} minimizes a weighted cumulative loss function:
\begin{equation}
   L(y, f(x ; \theta)) = \sum_i 
   \lambda_i \text{CE}(y, f_i(x ; \theta_i)),
\end{equation}
where $ 
\lambda_i > 0$ is the coefficient for the $i$-th ($i=1,\ldots, k$) classifier, and $\text{CE}(\cdot, \cdot)$ denotes the cross-entropy loss function. In MSDNet, all the $
\lambda_i$'s are simply set to 1.

This form of loss functions may lead to a \textit{gradient imbalance} issue due to the overlap of the subnetworks. Specifically, consider training a $k$-exit adaptive network using the sum of the cross-entropy losses of all the classifiers.
The backward graph can be described by a binary tree with depth $k$, where the gradients come from leave nodes and propagate from child nodes to father nodes. The gradient of the $i$-th block is contributed by the $i$-th node as well as the subsequent $(k-i)$ leave nodes:
\begin{equation}
   \nabla_{w_i}{L} = \sum_{i \leq j \leq k} 
   \lambda_i  \nabla_{w_i}{\text{CE}_j},
\end{equation}
where $w_i$ denotes the features at the $i$-th stage. 

From the above equation, it is easy to see that the total variance of $\nabla L$ can become very large as the gradients propagate backward. Formally, consider a situation that the gradients of the loss w.r.t. each $w_i$ are irrelevant, and the total variance of $\nabla_{w_i}{L}$ is computed by:
\begin{equation}
   \text{Var} (\nabla_{w_i}{L})  = {\sum_{i \leq j \leq k}{\lambda_j^2  \text{Var}(\nabla_{w_i} \text{CE}_j)}}.
\end{equation}

As the number of subnetworks increases, the variance of the gradient may grow overly large, leading to unstable training. Note that this issue can be fixed by averaging the cumulative loss, i.e., $\frac{1}{k}\sum_{i}{
   \lambda_i \text{CE}_i}$, but it tends to result in overly small gradients, which hinders the convergence.


To address this issue, we propose a \textbf{Gradient Equilibrium (GE)} method which re-normalizes the gradients at father nodes while maintaining the information flow in the forward procedure. The GE method consists of a series of \textit{gradient re-scaling} operations $R(\cdot; s)$ where $s$ is a scaling factor:
\begin{equation}
   R(x; s) = x; \ \ \ \nabla_x R(x; s) = {s}.
\end{equation}


To stabilize the backward procedure and resolve the gradient conflict, we propose to re-normalize the gradients in the following manner. For the $i$-th branch, we add two re-scaling modules for the gradients contributed by the current $i$-th classifier and the subsequent $(k-i)$ classifiers, setting their $s$ to $\frac{1}{k-i+1}$ and $\frac{k-i}{k-i+1}$, respectively. This ensures that the gradients have a bounded scale. To see this, we first calculate the gradient of $L_j$ w.r.t. $w_i$, with the re-scaling factors given above: 
\begin{equation}
\begin{aligned}
    \nabla_{w_i}^{\text{(GE)}} L_j 
    &=  \prod_{i \leq h < j} \frac{k-h}{k-h+1}  \!\times\! \frac{1}{k-j+1} \!\times\! \nabla_{w_i} L_j \\
    &= \frac{1}{k-i+1} \nabla_{w_i} L_j.
\end{aligned}
\end{equation}

For simplicity, we let $n \!=\! k\!-\!i\!+\!1$ and $X_j \!=\! \nabla_{w_i} L_{i+j-1}$. Then we have
\begin{equation}
    \begin{aligned}
        &\text{Var}\bigg(\sum_{j = i}^{k} \nabla_{w_i}^{\text{(GE)}} L_j\bigg) \\
        &= \text{Var}\bigg(\frac{1}{k-i+1} \sum_{j = i}^{k}\nabla_{w_i} L_j\bigg) \\
        &= \text{Var}\bigg(\frac{1}{n} \sum_{j=1}^n X_j\bigg)\\
        &= \frac{1}{n^2}\bigg(\sum_{j=1}^n\text{Var}(X_j) + \sum_{m\neq j}\text{Cov}(X_m, X_j)\bigg) \\
        &\leq \frac{1}{n^2}\bigg (\sum_{j=1}^n\text{Var}(X_j) + \sum_{m\neq j}\sqrt{\text{Var}(X_m)\text{Var}(X_j)}\bigg) \\
        &\leq \frac{1}{n^2} \bigg(n\max_l(\text{Var}(X_l)) + n(n-1)\max_l(\text{Var}(X_l))\bigg) \\
        &\leq \frac{2}{n^2} * n^2 * \max_l(\text{Var}(X_l)) \\
        &= 2\max_l(\text{Var}(X_l)) < \infty  
    \end{aligned}
\label{eq:summation}
\end{equation}

\subsection{Forward Knowledge Transfer}

In this and the following subsections, we aim to encourage collaboration among different classifiers in adaptive networks. In existing work, different exits are usually treated as independent models, except that their losses are simply summed up during training process. In fact, these classifiers heavily share parameters, and they are combined to solve the same task at test time, hinting that a collaborated learning process may significantly improve the training efficacy. Therefore, we propose two approaches to distill this insight into practical algorithms.

Our first approach is to promote forward knowledge transfer in adaptive networks. Specifically, we add a \emph{knowledge transfer path} between every two adjacent classifiers, to directly bypass the prediction at the $i$-th stage to the $(i+1)$-th classifier ($i=1,\ldots,k-1$). The knowledge transfer path may correspond to a tiny fully connected network or some functions without learning ability. Our experimental results show that even the simplest identity transform improves the performance of adaptive inference, where each classifier (except the very first one) can be considered as performing residual learning. Note that in this case, we discard the gradients along knowledge transform path in back propagation to prevent a classifier from being negatively affected by the latter ones.

Similar to the Knowledge Distillation algorithm \cite{hinton2015distilling}, we use the logits of the $i$-th classifier as the knowledge to facilitate the learning of its subsequent classifier, and we call the above approach \textbf{Inline Subnetwork Collaboration (ISC)}. 
Although being very simple, ISC consistently improves the performance of adaptive networks in our experiments.





\subsection{Backward Knowledge Transfer}

In the previous subsection, we introduce how to use the prediction from an early classifier to boost the performance for the latter classifiers. Here we introduce an approach to utilize the deepest classifier to help the learning of shallow classifiers. In a $k$-exit adaptive network, the last classifier usually achieves the best accuracy due to its highest capacity. This motivates us to adopt the knowledge distillation algorithm in the network. We call the approach \textbf{One-for-all Knowledge Distillation (OFA)}, as all the intermediate exits are supervised by the last classifier. 

In specific, the loss function for the $i$-th classifier consists of two parts weighted by a coefficient $\alpha$:

\begin{equation}
   L_i = \alpha  \text{CE}_i + (1-\alpha)  \text{KLD}_i,
\end{equation}
where $\text{CE}_i$ is the Cross-Entropy loss, and $\text{KLD}_i$ quantifies the alignment of \textit{soft} class probabilities between the teacher and student models using the Kullback Leibler divergence:
\begin{equation}
   \text{KLD}_i = -\sum_{c\in Y} p_k(c \mid x ; \theta, T)  \log{\frac{p_i(c \mid x ; \theta, T)}{p_k(c \mid x ; \theta, T)}}.
\end{equation}

%% file: exp.tex
\section{Experiments}

To demonstrate the effectiveness of our approach, we conducted extensive experiments on three representative image classification datasets, i.e., the CIFAR-10, CIFAR-100~\cite{cifar} and ILSVRC 2012 (ImageNet)~\cite{deng2009imagenet}. In addition, ablation studies are performed to analyze the three components of our method. All of our experiments are conducted on the multi-scale dense network (MSDNet) proposed in \cite{huang2018multi}, with the model re-implemented in PyTorch. Code to reproduce our results is avaliable at \url{https://github.com/kalviny/IMTA}. 

\paragraph{Datasets.}

The CIFAR-10 and CIFAR-100 datasets contain RGB images of size $32 \times 32$, corresponding to 10 and 100 classes, respectively. They both contain $50,000$ images for training and $10,000$ images for testing. Following \cite{huang2018multi}, we hold out $5,000$ training images as a validation set to search the confidence threshold for adaptive inference. We apply standard data augmentation schemes~\cite{resnet}: 1) images are zero-padded with $4$ pixels on each side, and then randomly cropped to produce $32 \times 32$ inputs; 2) images are horizontally flipped with probability $0.5$; 3) RGB channels are normalized by subtracting the corresponding channel mean and divided by their standard deviation.

The ImageNet dataset contains $1,000$ classes, with $1.2$ million training images and $50,000$ for testing \footnote{This subset is usually referred to as the validation set, as the true test set has not been made public. But in order to avoid confusion with the additional validation set we hold out from the training set, we view this subset as our test set.}. We hold out $50,000$ images from the training set as the validation set. We follow the practice in \cite{resnet,huang2017densely} for data augmentation at training time. At test time, images are firstly rescaled to $256 \times 256$ followed by a single $224 \times 224$ center crop, which are finally classified by the network.

\paragraph{Training Details.}

On the two CIFAR datasets, we optimize all models using stochastic gradient descent (SGD) with a mini-batch size $64$. 
We use Nesterov momentum with a momentum weight of $0.9$ without dampening, and a weight dacay of $10^{-4}$. 
The training is split into two phases. 
In phase I, the models are trained from scratch with Gradient Equilibrium for $300$ epochs, with an initial learning rate of $0.1$, which is further divided by a factor of $10$ after $150$ and $225$ epochs. 
In phase II, we start with the model obtained from phase I, and fine-tune only the last layer of each classifier with the proposed One-for-all Knowledge Distillation (OFA) and Inline Subnetwork Collaboration (ISC). This stage lasts for 180 epochs, with an initial learning rate of $0.1$ and divided by $10$ after $90$ and $135$ epochs, respectively. 
We apply the same training scheme to the ImageNet dataset, except that we increase the mini-batch size to 256, and all the models are trained for 90 epochs both in phase I and II with learning rate drops after 30 and 60 epochs.

\paragraph{Adaptive inference with MSDNet.} 
Following \cite{huang2018multi}, we evaluate our model in the adaptive inference setting. 
For a given input image, we forward it through the intermediate classifiers in an one-by-one manner. At each exit, we compare the prediction confidence, which is the highest softmax probability in our experiment, to a threshold, which is dependent on the given computational budget. If the current classifier is sufficiently confident about its prediction, i.e., the confidence value is greater than the threshold, the current prediction is used as the final prediction, and the latter classifiers are not evaluated. Otherwise, the subsequent classifier is evaluated, until a sufficient high confidence has been obtained, or the last classifier is evaluated. Intuitively,  ``easy'' examples are predicted by early classifiers, while only  ``hard'' examples are propagated through the latter classifiers of the network. In practice, most samples are relatively easy, thus this adaptive evaluation procedure can drastically improve the inference efficiency by saving computation on those large portion of ``easy'' samples in the dataset.

\begin{figure}[bpt]
   \begin{center}
     \includegraphics[width=0.97\linewidth]{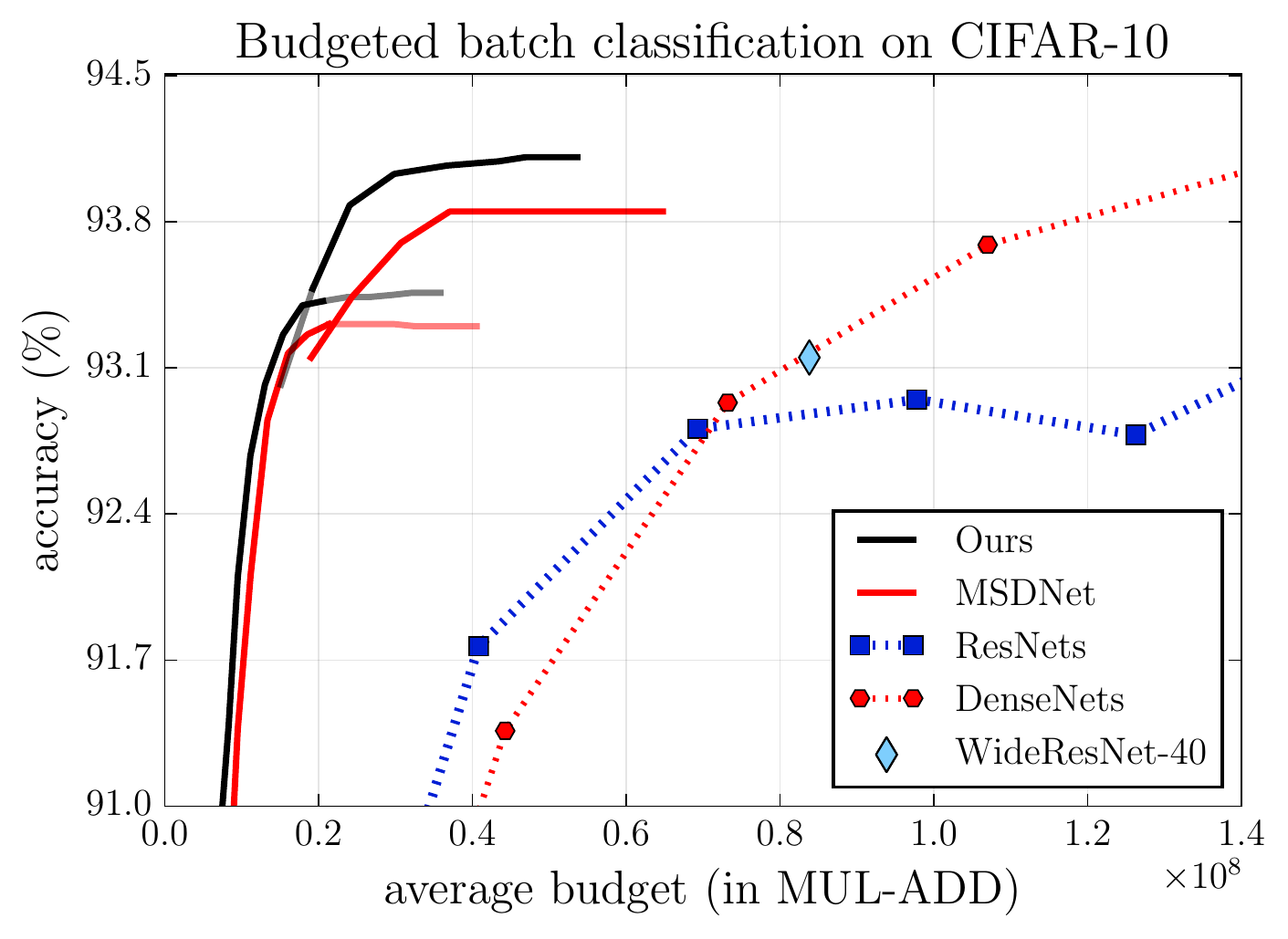}
   \end{center}
   \vspace{-2ex}
      \caption{Accuracy (top-1) of \textit{budgeted batch classification} as a function of average computational budget per image on the CIFAR-10.}
      \vspace{-2ex}
   \label{fig:budget_cifar10}
\end{figure}

\begin{figure}[bpt]
   \begin{center}
     \includegraphics[width=0.97\linewidth]{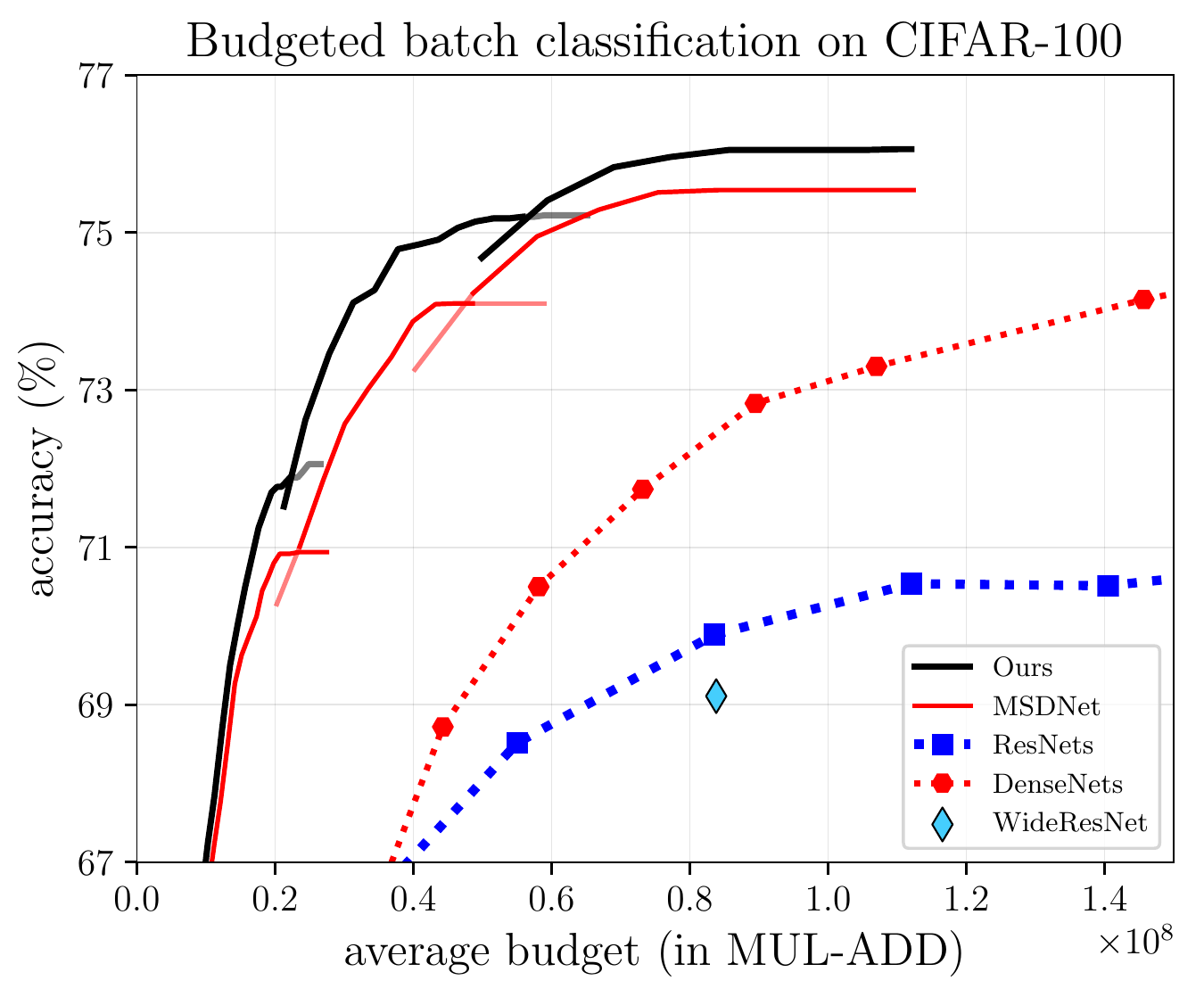}
   \end{center}
   \vspace{-2ex}
      \caption{Accuracy (top-1) of \textit{budgeted batch classification} as a function of average computational budget per image on the CIFAR-100.}
      \vspace{-2ex}
   \label{fig:budget_cifar100}
\end{figure}

\begin{figure}[bpt]
   \centering   
   \subfigure[]{\includegraphics[width=0.49\linewidth]{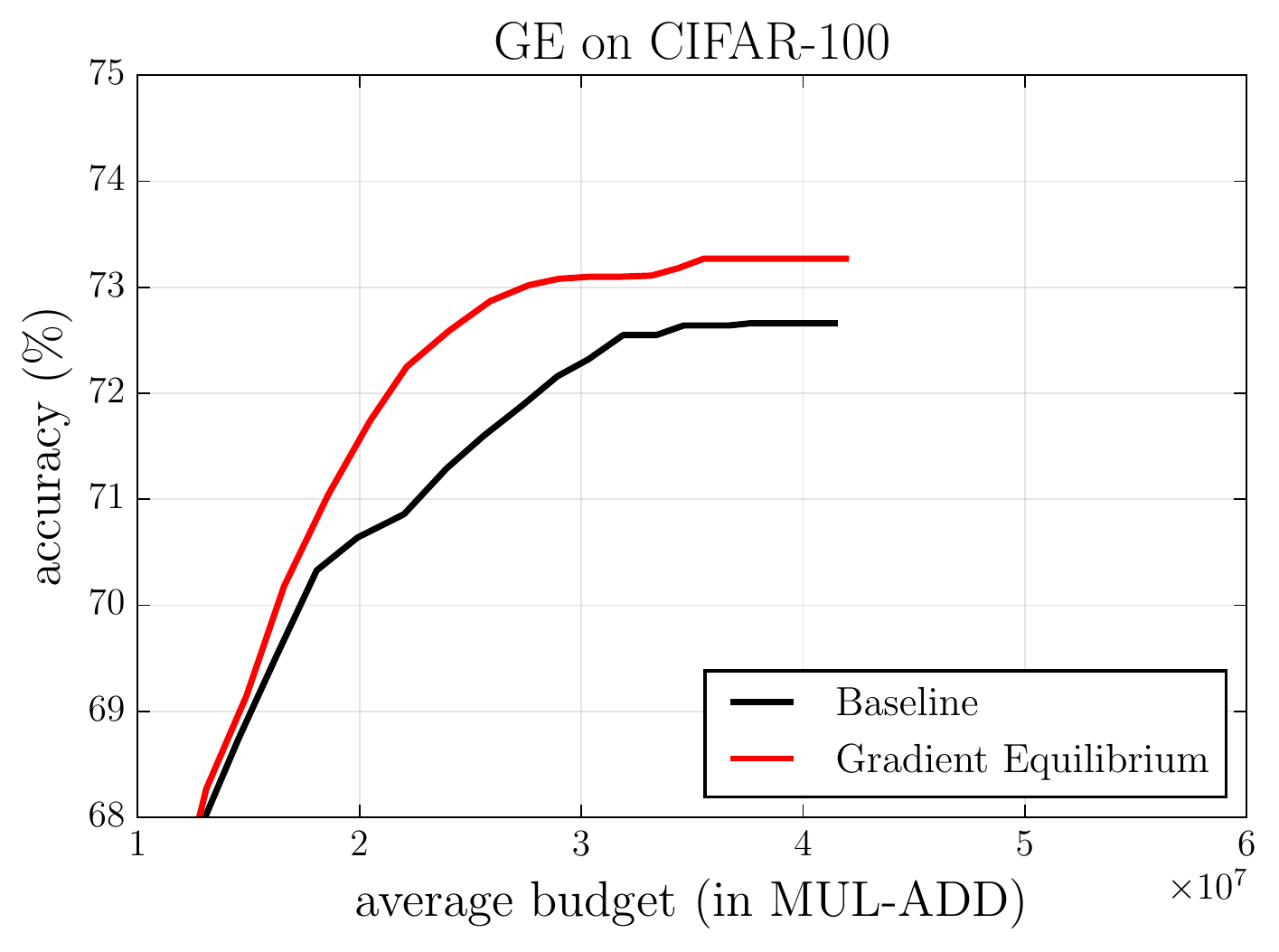}}
   \subfigure[]{\includegraphics[width=0.49\linewidth]{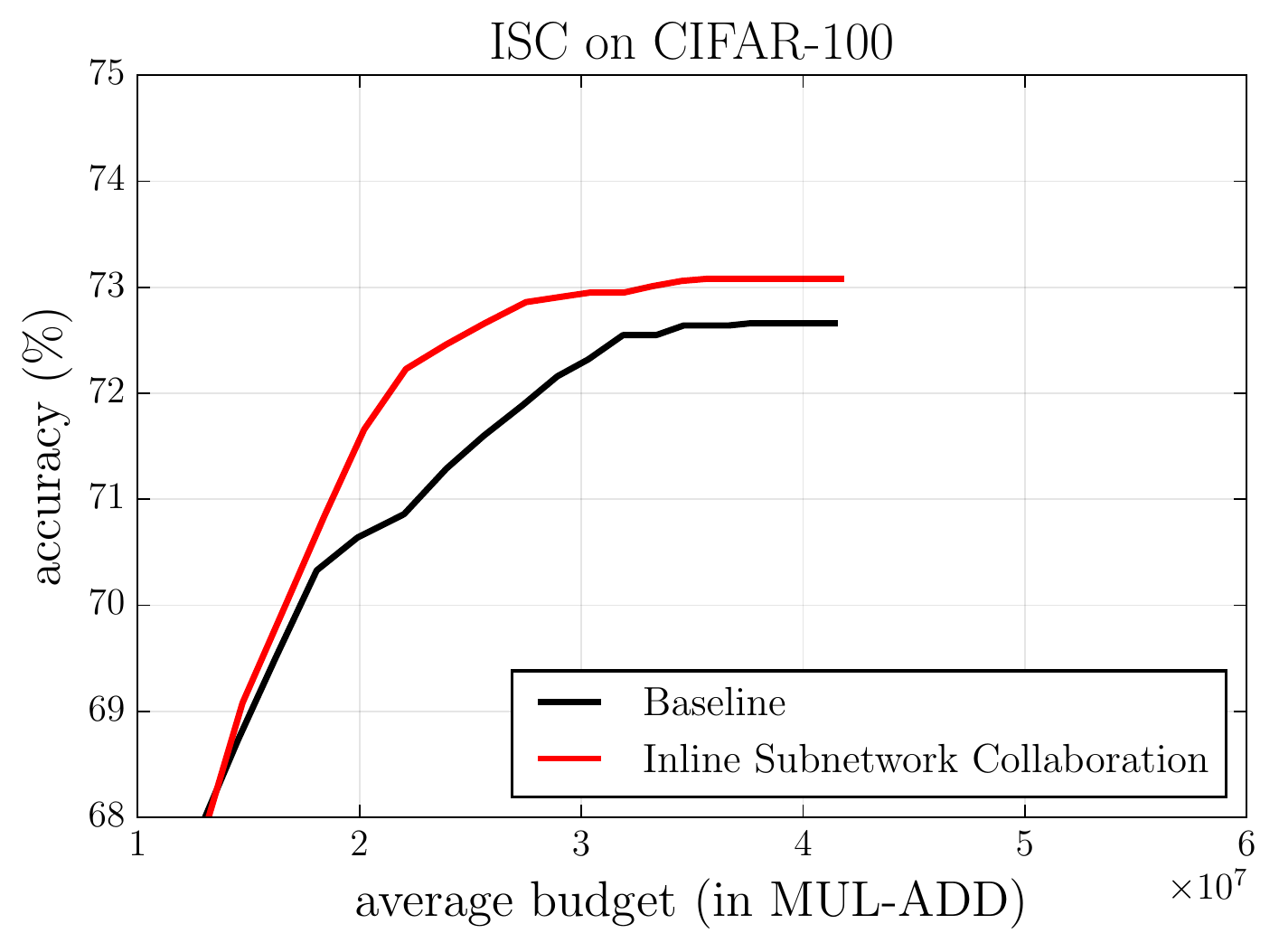}}
   \subfigure[]{\includegraphics[width=0.49\linewidth]{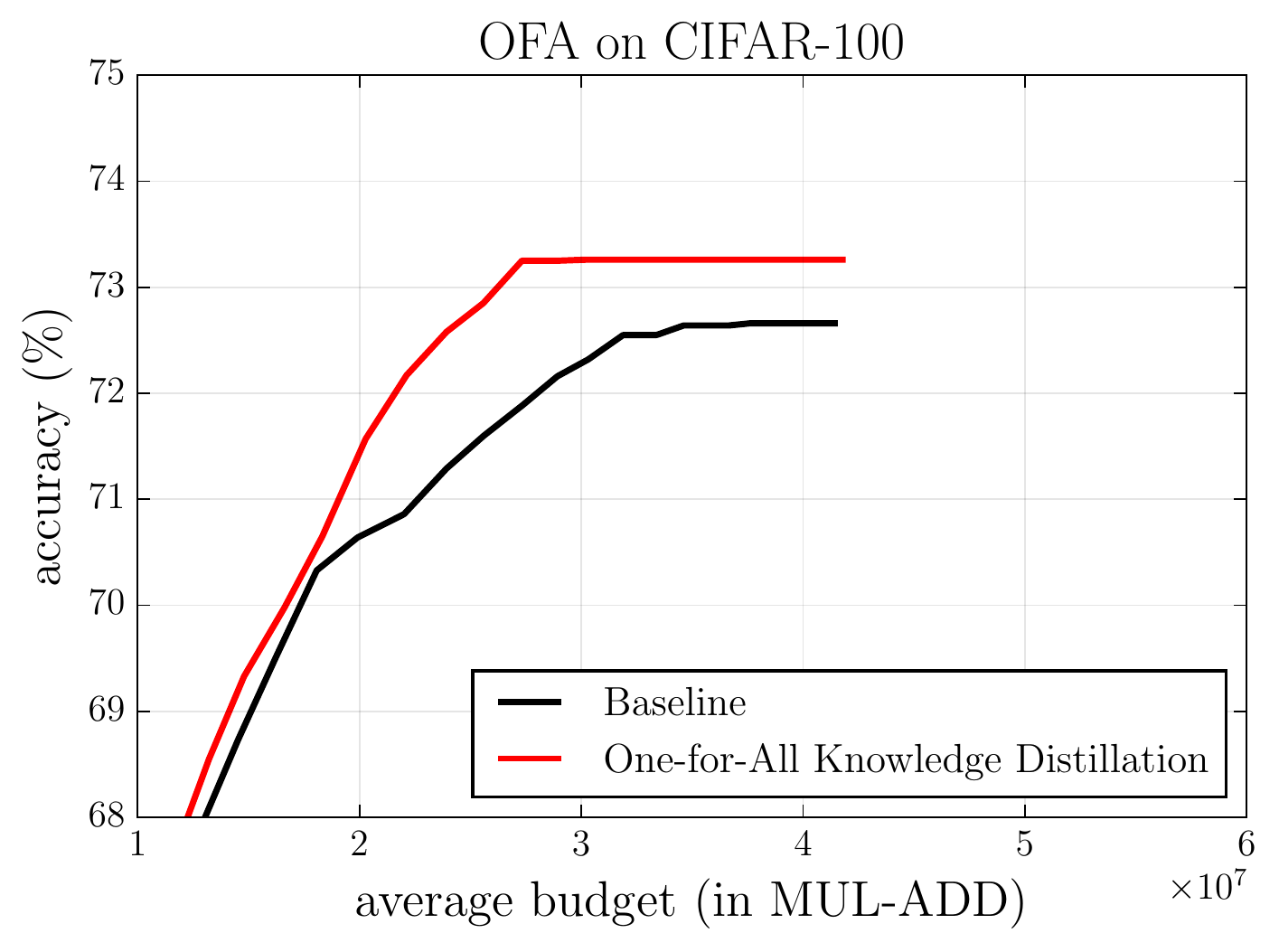}}
   \subfigure[]{\includegraphics[width=0.49\linewidth]{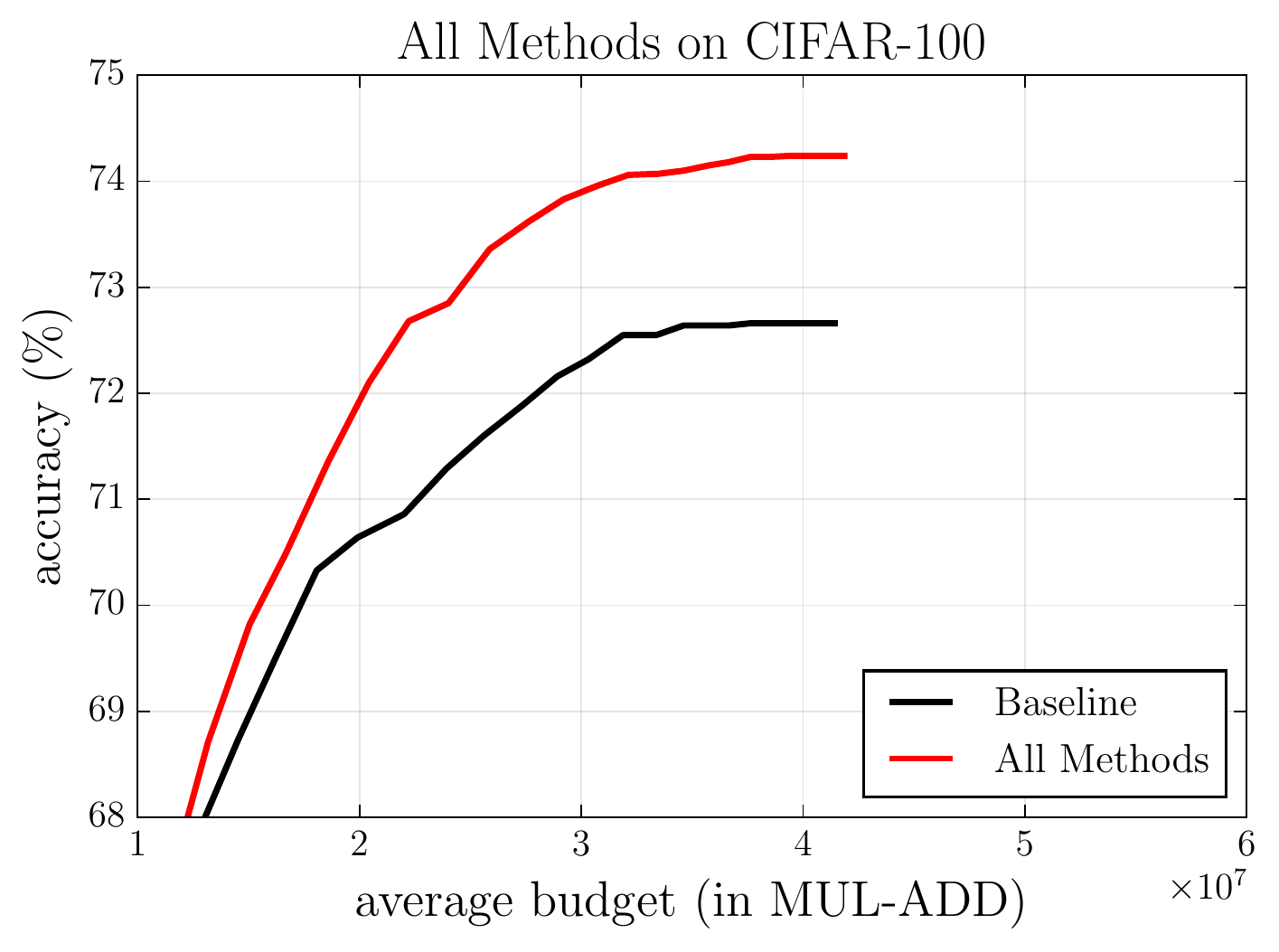}}
   \vspace{-1ex}
   \caption{Ablation results produced by integrating \textbf{GE, ISC, OFA} on CIFAR-100.}
   \vspace{-2ex}
   \label{fig:abaltion_budget}
\end{figure}

\paragraph{Compared Models.}
To verify the effectiveness of our approach for adaptive inference, we mainly compare with the following baseline models.
\begin{itemize}
   \item{MSDNet~\cite{huang2018multi}}. As our proposed learning strategies are also performed on MSDNet, it serves as a direct baseline for our experiments.  
   \item{ResNet~\cite{resnet} and DenseNet~\cite{huang2017densely}}. We also compare our approach with ResNets and DenseNets. We do not perform adaptive inference on these models, since they are not designed for this purpose, and are shown to yield inferior results compared to MSDNet \cite{huang2018multi}.

\end{itemize}

As we mainly focus on the training strategy of adaptive CNNs, we do not compare with efficient models with more advance architecure designs, such as the MobileNet \cite{howard2017mobilenets}, ShuffleNet \cite{zhang2018shufflenet} and NASNet \cite{zoph2018learning}. As discussed earlier in the paper, the architecture innovations for these models, like the depth separable convolutions, are orthogonal to adaptive inference methods, and in principle, they may benefit the adaptive models as well. To focus on the adaptive learning setting, we leave investigations on this direction for future work.


\subsection{Evaluation on CIFAR}

\paragraph{Baselines.}

On the two CIFAR datasets, following the MSDNet we train networks with three-scale features, \ie, $32 \times 32$, $16 \times 16$, $8 \times 8$. To better evaluate our approaches, on CIFAR-10, the MSDNets are with \textit{\{6, 8\}} exits and the depths are \textit{\{21, 36\}}. On CIFAR-100, we train MSDNets with \textit{\{4, 5, 6, 8\}} exits and the depths are \textit{\{10, 15, 21, 36\}} respectively.  In this setting, we also compare with the original baseline MSDNet, four ResNets with different depths, and six DensNets with varying depths.




 \begin{table}
   \begin{center}
   \begin{tabular}{ c c c c }
       \hline
           Model & \tabincell{c}{Params \\ $(\times 10^6)$} & \tabincell{c}{Inference MADDs \\ $\times 10^6$} & \tabincell{c}{Accuracy \\ (Top-1)} \\
           \hline 
           Hydra-Res-d1 & 1.28 & 52 & 65.81 \\
           Hydra-Res-d2 & 2.86 & 118 & 71.24 \\
           Hydra-Res-d3 & 4.43 & 184 & 72.30 \\
           Hydra-Res-d4 & 6.01 & 251 & 73.35 \\
           Hydra-Res-d5 & 7.59 & 317 & 73.84 \\
           Hydra-Res-d6 & 9.17 & 383 & 74.29 \\
           Hydra-Res-d7 & 10.74  & 449 & 74.71 \\
           Hydra-Res-d9 & 13.90 & 581 & 75.26 \\
           \hline 
           MSDNet-Exit1 & 0.3 & 6.86 & 64.1 \\
           MSDNet-Exit2 & 0.65 & 14.35 & 67.46 \\
           MSDNet-Exit3 & 1.11 & 27.29 & 70.34 \\
           MSDNet-Exit4 & 1.73 & 48.45 & 72.38 \\
           MSDNet-Exit5 & 2.38 & 76.43 & 73.06 \\
           MSDNet-Exit6 & 3.05 & 108.9 & 73.81 \\
           MSDNet-Exit7 & 4.0 & 137.3 & 73.89 \\
           \hline
           Ours-Exit1 & 0.3 & 6.86 & 64.00 \\
           Ours-Exit2 & 0.65 & 14.35 & 68.41 \\
           Ours-Exit3 & 1.11 & 27.29 & 71.86 \\
           Ours-Exit4 & 1.73 & 48.45 & 73.50 \\
           Ours-Exit5 & 2.38 & 76.43 & 74.46 \\
           Ours-Exit6 & 3.05 & 108.9 & 75.39 \\
           Ours-Exit7 & 4.0 & 137.3 & 75.96 \\
       \hline
   \end{tabular}
   \end{center}
   \caption{Classification accuracy of individual classifiers on CIFAR-100.} 
   \label{tab:hydra}
\end{table}

\begin{table*}[bpt]
   \begin{center}
   \begin{tabular}{ c c c|c|c|c|c|c }
       \multicolumn{3}{c|}{Method} & \multicolumn{5}{c}{Accuracy @TOP1} \\
       \hline
           GE & ISC & OFA & E-1 & E-2 & E-3 & E-4 & E-5 \\
       \hline
          - & - & - & 60.09 & 63.73 & 67.89 & 70.48 & 71.81 \\
       \checkmark &  &  & 60.35 & 64.38 & 68.72 & 70.65 & 71.94 \\
        & \checkmark &  & 60.19 & 64.72 & 68.07 & 70.94 & 73.28 \\
        &  & \checkmark & 60.39 & 64.20 & 68.10 & 70.65 & 71.85 \\ 
        \checkmark & \checkmark & \checkmark & \textbf{60.78} & \textbf{65.54} & \textbf{69.98} & \textbf{72.27} & \textbf{73.45} \\
   \end{tabular}
   \end{center}
   \caption{Accuracy at different exits on CIFAR-100. The results produced by integrating \textbf{GE, ISC, OFA}.} 
   \label{tab:abaltion_cifar}
   \end{table*}

\begin{table*}[bpt]
   \begin{center}
   \begin{tabular}{ c c c|c|c|c|c|c }
         \multicolumn{3}{c|}{Method} & \multicolumn{5}{c}{Accuracy @TOP1} \\
         \hline
            GE & ISC & OFA & E-1 & E-2 & E-3 & E-4 & E-5 \\
         \hline
            - & - & - & 56.64 & 65.14 & 68.42 & 69.77 & 71.34 \\
         \checkmark &  &  & 57.08 & 65.29 & 69.08 & 70.55 & 72.14 \\
         & \checkmark &  & 57.03 & 66.2 & 69.73 & 71.15 & 71.65 \\
         &  & \checkmark & 57.15 & 65.77 & 68.87 & 70.23 & 71.39 \\ 
         \checkmark & \checkmark & \checkmark & \textbf{57.28} & \textbf{66.22} & \textbf{70.24} & \textbf{71.71} & \textbf{72.43} \\
   \end{tabular}
   \end{center}
   \vspace{-1ex}
   \caption{Accuracy at different exits on ImageNet. The results produced by integrating \textbf{GE, ISC, OFA}.}
   \vspace{-2ex}
   \label{tab:ablation_imagenet}
\end{table*}
   

\paragraph{Results on CIFAR.}
The evaluation results on CIFAR-10 and CIFAR-100 are shown in Figure ~\ref{fig:budget_cifar10} and Figure ~\ref{fig:budget_cifar100}, respectively. The results for baseline MSDNet are plotted by red curves (corresponding to three MSDNets with different sizes), and the results for our proposed training strategy are shown by black curves. As shown in the figures, MSDNet trained with our proposed training strategy clearly achieves better accuracy than the baseline MSDNet under the same time budget.
Moreover, the improvement increases as we have more time budgets. 
For example, with $1 \times 10^8$ FLOPS, our training strategy improves MSDNet baseline by more than $0.5\%$ in terms of top-1 accuracy.  
This demonstrates that our training strategy can facilitate training of deeper layers.  
Besides, compared with ResNets and DenseNets, adaptive inference based approaches (MSDNet with and without our training strategies) perform significantly better with the same amount of computation. For instance, to achieve the same accuracy on CIFAR-100 (Figure~\ref{fig:budget_cifar100}), our approach requires half the amount of computation as DenseNet and 1/3 of the computation as ResNet.

In Table~\ref{tab:hydra}, we report the classification accuracy of all the individual classifiers of our model, and compare it with MSDNet as well as the recently proposed HydraNets \cite{teja2018hydranets}. We can observe that the results of each individual classifier of our network are competitive with state-of-the-art models.

\subsection{Evaluation on ImageNet}

\begin{figure}[bpt]
    \begin{center}
      \includegraphics[width=1.0\linewidth]{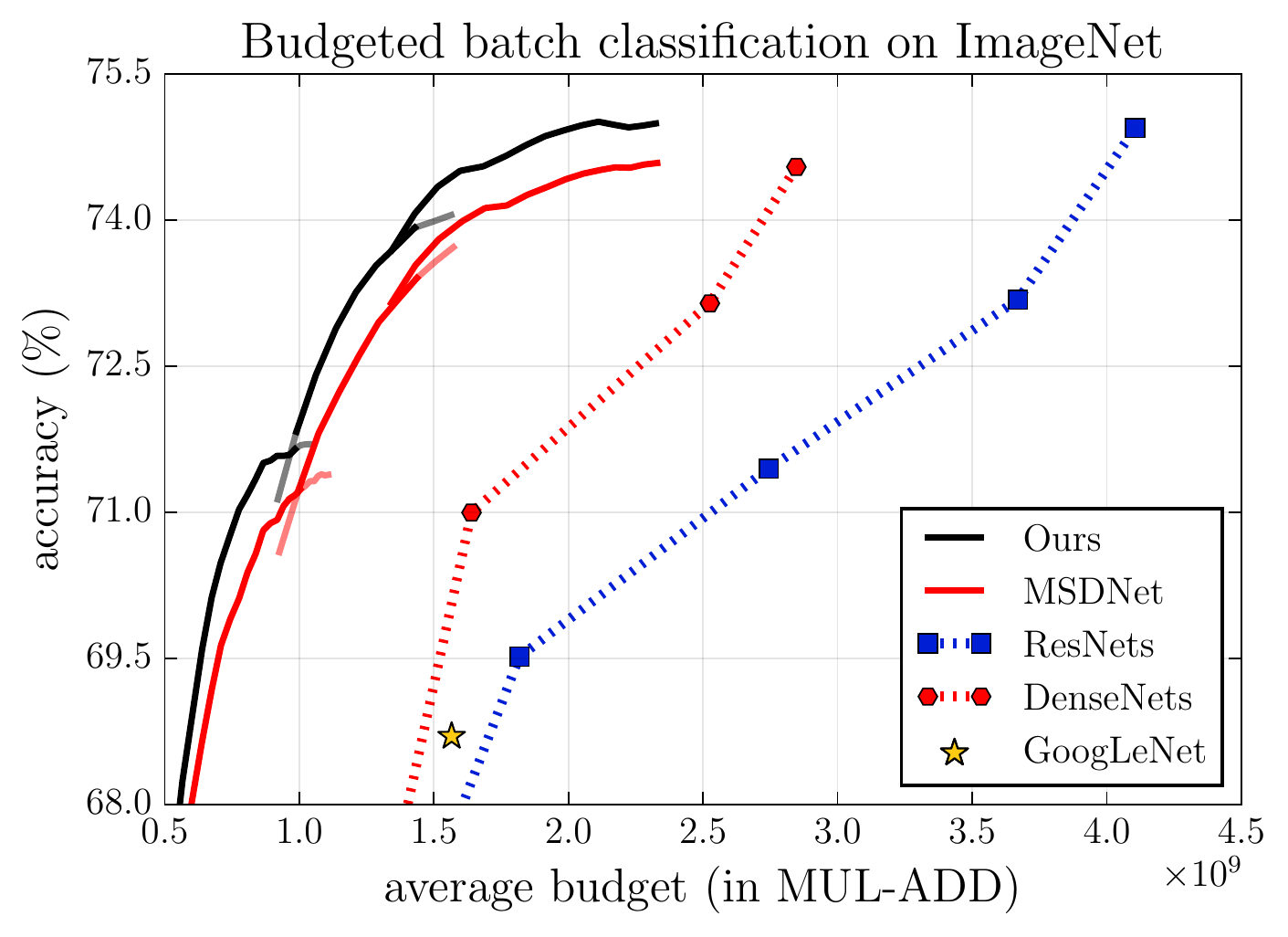}
    \end{center}
    \vspace{-1ex}
       \caption{Top-1 accuracy of \textit{budgeted batch classification} as a function of average computational budget per image on ImageNet.}
       \vspace{-2ex}
    \label{fig:budget_imagenet}
\end{figure}

\paragraph{Baselines.}

On ImageNet, we use the four-scale MSDNet, \ie, $56\times 56$, $28 \times 28$, $14 \times 14$, $7 \times 7$. Each of the MSDNet has five classifiers inserted at different depths.
Specifically, the $i^{th}$ classifier is attached in the $(t \times i + 3)^{th}$ layer where $i \in \{1, \dots ,5\}$, $t \in \{4, 6, 7\}$ is the step for each network block. 
We also compare with other competitive approaches, ResNet~\cite{resnet}, DenseNet~\cite{huang2017densely} and GoogleNet~\cite{szegedy2015going}.



\paragraph{Results on ImageNet.} The results on ImageNet are shown in Figure \ref{fig:budget_imagenet}. 
One can observe that our method with dynamic evaluation built on the top of MSDNets consistently surpass the baseline network. With $1 \times 10^8$ FLOPS computation budget, we improve the baseline method by around $0.5\%$ in terms of top-1 accuracy. Again, for the same MSDNet architecture, our method improves the baseline by a larger margin as more allowed computational budgets. This further verifies that our proposed method is effective for deeper adaptive networks.
Moreover, with the same FLOPs, our method is more accurate and efficient than the models of ResNets and DenseNets. For instance, with around $1\times 10^8$ FLOPs, our approach outperforms the ResNet and DensNet by more than $6\%$.

\subsection{Ablation Study}
To investigate the effectiveness of the individual modules of the proposed approach, i.e., \textbf{GE, ISC} and \textbf{OFA}, we conduct ablative analysis on the CIFAR-100. We set the baseline MSDNets on CIFAR-100 with three scales and five classifiers and on ImageNet we use MSDNet with four scales and five classifiers. Quantitive results are shown in Table \ref{tab:abaltion_cifar} and Table \ref{tab:ablation_imagenet}. Our full model consistently improves the accuracy on both CIFAR-100 and ImageNet. For stages two to four, our full model improves the baseline method by more than $1\%$ on CIFAR-100 dataset, and surpasses the baseline by more than $0.5\%$. All the strategies consistently improve the performance on both datasets.

\begin{figure}[bpt]
   \begin{center}
     \includegraphics[width=1.0\linewidth]{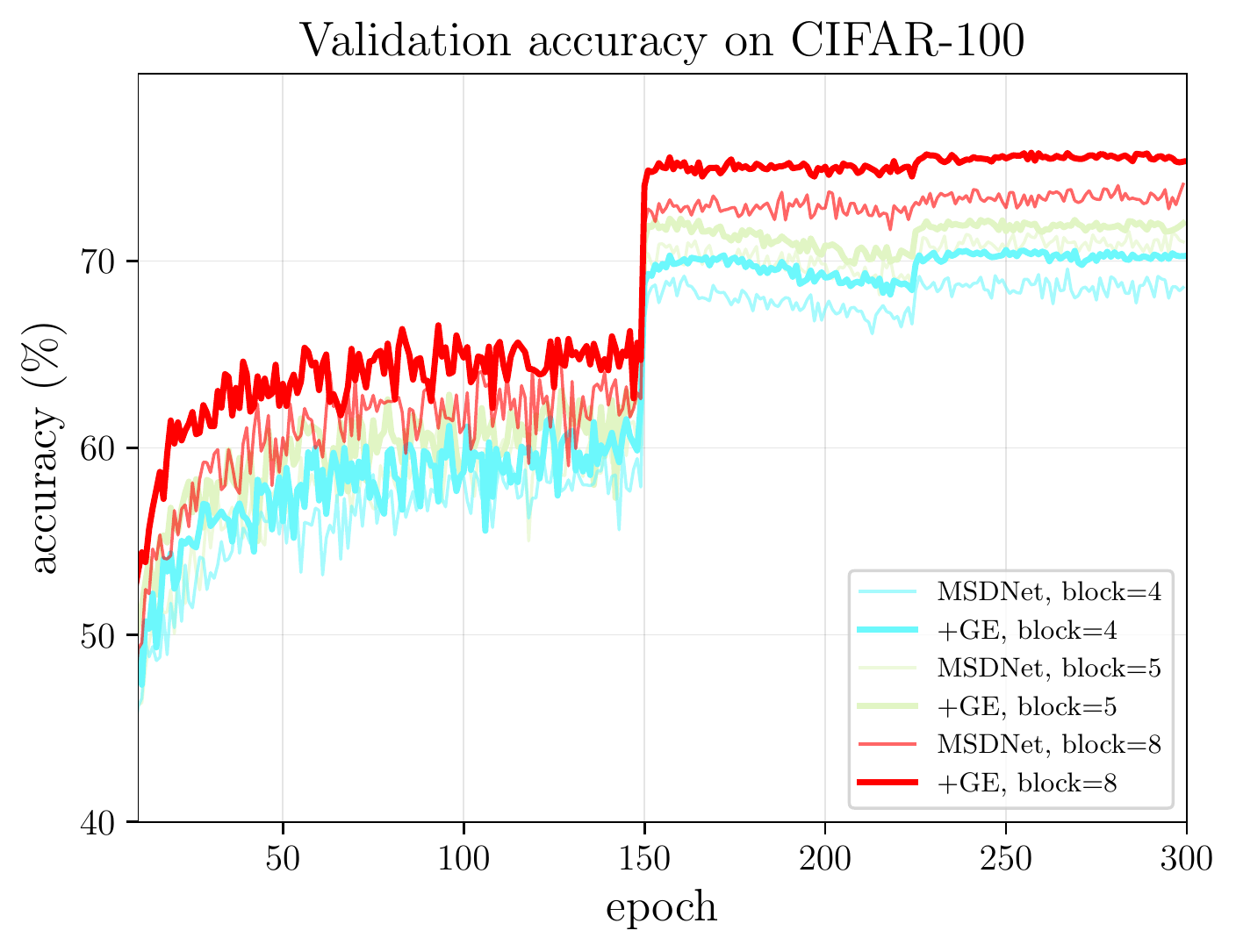}
   \end{center}
   \vspace{-1ex}
      \caption{\textbf{Validation accuracy} on CIFAR-100 at different epochs. We plot the results of three different depth of networks with 4, 6, 8 exits respectively. The losses of the models trained with \textbf{GE} are consistently lower than the baseline models.}
      \vspace{-2ex}
\label{fig:err_epoch}
\end{figure}

\paragraph{Gradient Equilibrium.}
To further evaluate the effectiveness of \textit{Gradient Equilibrium}, we compare different MSDNet architecture trained with \textit{Gradient Equilibrium} and with the baseline MSDNet.
Validation accuracy in both settings are shown in Figure~\ref{fig:err_epoch}.
With \textit{Gradient Equilibrium}, the validation accuracy is consistently higher and the training procedure is more stable than the baseline algorithm for all the compared architectures. This demonstrates \textit{Gradient Equilibrium} can help stabilize the training process and at the same time improve the accuracy which is also shown in Table~\ref{tab:abaltion_cifar}(line 1 vs line 2) and Table~\ref{tab:ablation_imagenet}.

\begin{figure}[bpt]
   \begin{center}
     \includegraphics[width=1.0\linewidth]{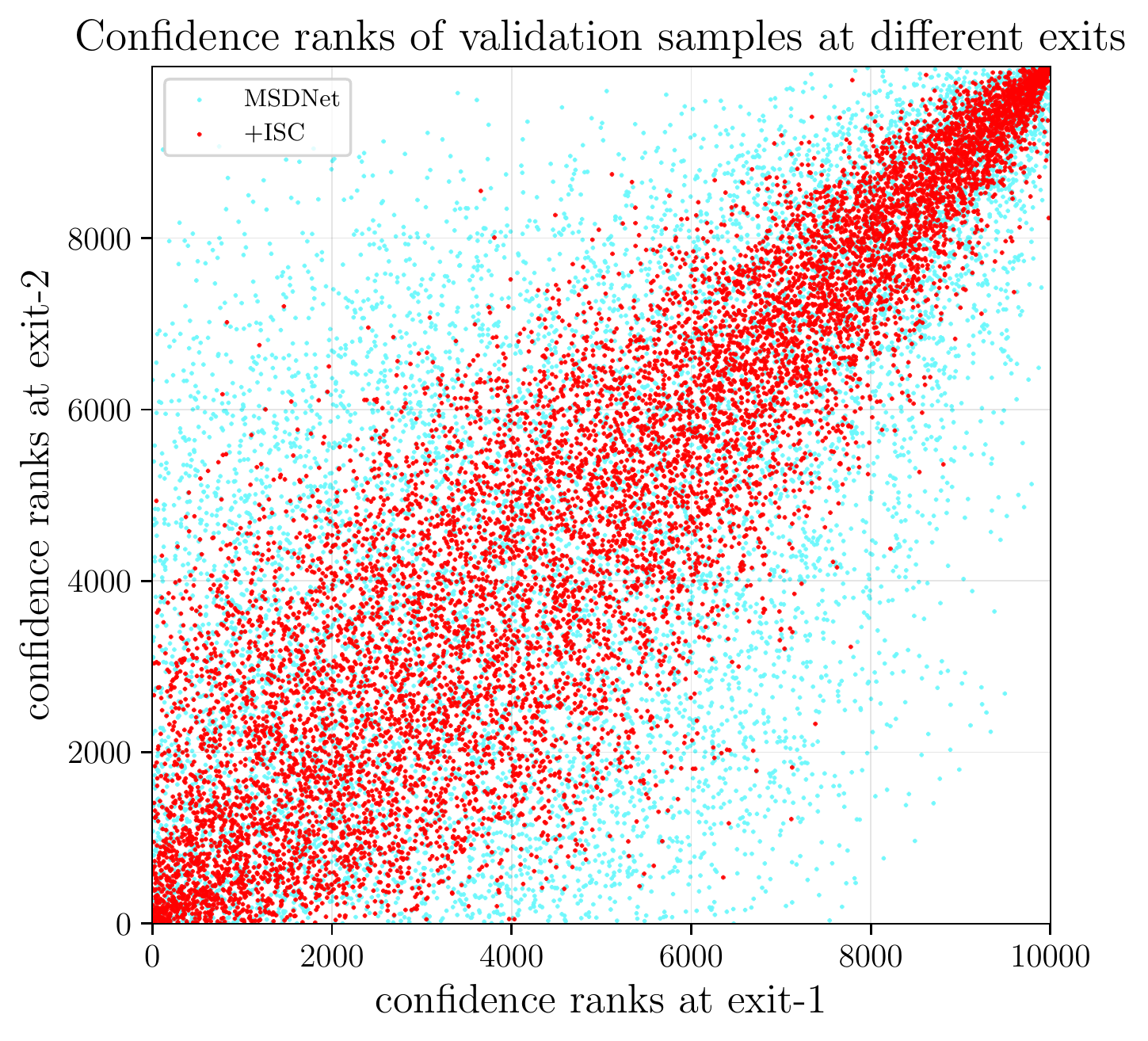}
   \end{center}
   \vspace{-1ex}
   \caption{The distribution of \textit{confidence scores} at different exits. In order to investigate the ISC's effects to different classifiers, we compare the ranks of each sample on different exits. It's obvious that the distribution is more consistent with ISC, i.e. more closer to the identity mapping, which indicates ISC helps the collaboration among exits. }
   \vspace{-2ex}
   \label{fig:conf_rank}
\end{figure}

\paragraph{Inline Subnetwork Collaboration.}
Inline Subnetwork Collaboration (ISC) can consistently improve the performance as shown in Table \ref{tab:abaltion_cifar} (line 1 vs line 3) on CIFAR-100 dataset. Deeper layers typically benefit more from ISC, \eg, exit 5 (E-5) improves by more than $1.4\%$ in Top-1 accuracy. This might be explained by that deeper layer can acquire more information from other classifiers with our inline subnetwork collaboration module.
In Figure \ref{fig:conf_rank}, we plot the confidence rank of all the validation samples before and after applying ISC. One can observe a clear trend that the red dots, which corresponding to the results with ISC, is more concentrated than the blue dots, which corresponding to the results without ISC. This demonstrates that with ISC, the consistency between the rank of samples at different exits (exit-1 and exit-2 here) has significantly increased, and partially explains the effectiveness of ISC.

\paragraph{One-for-all Knowledge Distillation.}
Quantitative improvements with the OFA strategy are shown in Table.~\ref{tab:abaltion_cifar} (line 1 vs line 4). Lower layers tend to gain larger improvement, showing that they benefited from the supervision from the deepest classifier. These results further show that knowledge distillation is indeed effective for the adaptive network to exploit its own prediction.

%% file: conclusion.tex

\section{Conclusion}
In this paper, we have presented three techniques to improve the training of adaptive neural network with multiple exits. On one hand, a Gradient Equilibrium (GE) approach is proposed to stabilize the training procedure and resolve the conflict of learning objectives of different classifiers. One the other hand, we have introduced two techniques to strengthen collaboration among classifiers. Although being simple, the proposed techniques have shown its effectiveness on a number of image recognition datasets, and significantly improved the efficiency of the recently proposed MSDNet. Future research may focus on extending our results to other types of adaptive networks, e.g., spatially adaptive networks \cite{figurnov2016spatially}, or applying them to other computer vision tasks, such as object detection, semantic segmentation and image generation. 

%% file: main.bbl
\begin{thebibliography}{10}\itemsep=-1pt

\bibitem{ba2014deep}
Jimmy Ba and Rich Caruana.
\newblock Do deep nets really need to be deep?
\newblock In {\em NIPS}, 2014.

\bibitem{bolukbasi2017adaptive}
Tolga Bolukbasi, Joseph Wang, Ofer Dekel, and Venkatesh Saligrama.
\newblock Adaptive neural networks for fast test-time prediction.
\newblock In {\em ICML}, 2017.

\bibitem{bucilu2006model}
Cristian Bucilua, Rich Caruana, and Alexandru Niculescu-Mizil.
\newblock Model compression.
\newblock In {\em ACM SIGKDD}, 2006.

\bibitem{chen2016compressing}
Wenlin Chen, James Wilson, Stephen Tyree, Kilian~Q Weinberger, and Yixin Chen.
\newblock Compressing convolutional neural networks in the frequency domain.
\newblock In {\em ACM SIGKDD}, 2016.

\bibitem{deng2009imagenet}
Jia Deng, Wei Dong, Richard Socher, Li-Jia Li, Kai Li, and Li Fei-Fei.
\newblock Imagenet: A large-scale hierarchical image database.
\newblock In {\em CVPR}, 2009.

\bibitem{figurnov2016spatially}
Michael Figurnov, Maxwell~D Collins, Yukun Zhu, Li Zhang, Jonathan Huang,
  Dmitry Vetrov, and Ruslan Salakhutdinov.
\newblock Spatially adaptive computation time for residual networks.
\newblock {\em arXiv preprint arXiv:1612.02297}, 2016.

\bibitem{graves2016adaptive}
Alex Graves.
\newblock Adaptive computation time for recurrent neural networks.
\newblock {\em arXiv preprint arXiv:1603.08983}, 2016.

\bibitem{han2015deep}
Song Han, Huizi Mao, and William~J Dally.
\newblock Deep compression: Compressing deep neural networks with pruning,
  trained quantization and huffman coding.
\newblock In {\em ICLR}, 2016.

\bibitem{he2017mask}
Kaiming He, Georgia Gkioxari, Piotr Doll{\'a}r, and Ross Girshick.
\newblock Mask r-cnn.
\newblock In {\em ICCV}, 2017.

\bibitem{resnet}
Kaiming He, Xiangyu Zhang, Shaoqing Ren, and Jian Sun.
\newblock Deep residual learning for image recognition.
\newblock In {\em CVPR}, 2016.

\bibitem{he2018amc}
Yihui He, Ji Lin, Zhijian Liu, Hanrui Wang, Li-Jia Li, and Song Han.
\newblock Amc: Automl for model compression and acceleration on mobile devices.
\newblock In {\em ECCV}, 2018.

\bibitem{hinton2015distilling}
Geoffrey Hinton, Oriol Vinyals, and Jeff Dean.
\newblock Distilling the knowledge in a neural network.
\newblock In {\em NIPS Deep Learning Workshop}, 2014.

\bibitem{howard2017mobilenets}
Andrew~G Howard, Menglong Zhu, Bo Chen, Dmitry Kalenichenko, Weijun Wang,
  Tobias Weyand, Marco Andreetto, and Hartwig Adam.
\newblock Mobilenets: Efficient convolutional neural networks for mobile vision
  applications.
\newblock {\em arXiv preprint arXiv:1704.04861}, 2017.

\bibitem{huang2018multi}
Gao Huang, Danlu Chen, Tianhong Li, Felix Wu, Laurens van~der Maaten, and
  Kilian~Q Weinberger.
\newblock Multi-scale dense networks for resource efficient image
  classification.
\newblock In {\em ICLR}, 2018.

\bibitem{huang2018condensenet}
Gao Huang, Shichen Liu, Laurens Van~der Maaten, and Kilian~Q Weinberger.
\newblock Condensenet: An efficient densenet using learned group convolutions.
\newblock In {\em CVPR}, 2018.

\bibitem{huang2017densely}
Gao Huang, Zhuang Liu, Kilian~Q Weinberger, and Laurens van~der Maaten.
\newblock Densely connected convolutional networks.
\newblock In {\em CVPR}, 2017.

\bibitem{hubara2016binarized}
Itay Hubara, Matthieu Courbariaux, Daniel Soudry, Ran El-Yaniv, and Yoshua
  Bengio.
\newblock Binarized neural networks.
\newblock In {\em NIPS}, 2016.

\bibitem{kang2017incorporating}
Di Kang, Debarun Dhar, and Antoni Chan.
\newblock Incorporating side information by adaptive convolution.
\newblock In {\em NIPS}, 2017.

\bibitem{kong2018pixel}
Shu Kong and Charless Fowlkes.
\newblock Pixel-wise attentional gating for parsimonious pixel labeling.
\newblock {\em arXiv preprint arXiv:1805.01556}, 2018.

\bibitem{cifar}
Alex Krizhevsky and Geoffrey Hinton.
\newblock Learning multiple layers of features from tiny images.
\newblock {\em Tech Report}, 2009.

\bibitem{alexnet}
Alex Krizhevsky, Ilya Sutskever, and Geoffrey~E Hinton.
\newblock Imagenet classification with deep convolutional neural networks.
\newblock In {\em NIPS}, 2012.

\bibitem{lan2018knowledge}
Xu Lan, Xiatian Zhu, and Shaogang Gong.
\newblock Knowledge distillation by on-the-fly native ensemble.
\newblock {\em arXiv preprint arXiv:1806.04606}, 2018.

\bibitem{li2016pruning}
Hao Li, Asim Kadav, Igor Durdanovic, Hanan Samet, and Hans~Peter Graf.
\newblock Pruning filters for efficient convnets.
\newblock In {\em ICLR}, 2017.

\bibitem{li2017dynamic}
Zhichao Li, Yi Yang, Xiao Liu, Feng Zhou, Shilei Wen, and Wei Xu.
\newblock Dynamic computational time for visual attention.
\newblock In {\em ICCV}, 2017.

\bibitem{lin2017runtime}
Ji Lin, Yongming Rao, Jiwen Lu, and Jie Zhou.
\newblock Runtime neural pruning.
\newblock In {\em NIPS}, 2017.

\bibitem{liu2017learning}
Zhuang Liu, Jianguo Li, Zhiqiang Shen, Gao Huang, Shoumeng Yan, and Changshui
  Zhang.
\newblock Learning efficient convolutional networks through network slimming.
\newblock In {\em ICCV}, 2017.

\bibitem{fcn}
Jonathan Long, Evan Shelhamer, and Trevor Darrell.
\newblock Fully convolutional networks for semantic segmentation.
\newblock In {\em CVPR}, 2015.

\bibitem{ma2018shufflenet}
Ningning Ma, Xiangyu Zhang, Hai-Tao Zheng, and Jian Sun.
\newblock Shufflenet v2: Practical guidelines for efficient cnn architecture
  design.
\newblock In {\em ECCV}, 2018.

\bibitem{mcintosh2018recurrent}
Lane McIntosh, Niru Maheswaranathan, David Sussillo, and Jonathon Shlens.
\newblock Recurrent segmentation for variable computational budgets.
\newblock In {\em CVPR Workshops}, 2018.

\bibitem{sandler2018mobilenetv2}
Mark Sandler, Andrew Howard, Menglong Zhu, Andrey Zhmoginov, and Liang-Chieh
  Chen.
\newblock Mobilenetv2: Inverted residuals and linear bottlenecks.
\newblock In {\em CVPR}, 2018.

\bibitem{song2018collaborative}
Guocong Song and Wei Chai.
\newblock Collaborative learning for deep neural networks.
\newblock In {\em NIPS}, 2018.

\bibitem{szegedy2015going}
Christian Szegedy, Wei Liu, Yangqing Jia, Pierre Sermanet, Scott Reed, Dragomir
  Anguelov, Dumitru Erhan, Vincent Vanhoucke, and Andrew Rabinovich.
\newblock Going deeper with convolutions.
\newblock In {\em CVPR}, 2015.

\bibitem{teerapittayanon2016branchynet}
Surat Teerapittayanon, Bradley McDanel, and HT Kung.
\newblock Branchynet: Fast inference via early exiting from deep neural
  networks.
\newblock In {\em ICPR}, 2016.

\bibitem{teja2018hydranets}
Ravi Teja~Mullapudi, William~R Mark, Noam Shazeer, and Kayvon Fatahalian.
\newblock Hydranets: Specialized dynamic architectures for efficient inference.
\newblock In {\em CVPR}, 2018.

\bibitem{veit17conditionalsimilarity}
Andreas Veit and Serge Belongie.
\newblock Convolutional networks with adaptive inference graphs.
\newblock In {\em ECCV}, 2018.

\bibitem{wang2018skipnet}
Xin Wang, Fisher Yu, Zi-Yi Dou, Trevor Darrell, and Joseph~E Gonzalez.
\newblock Skipnet: Learning dynamic routing in convolutional networks.
\newblock In {\em ECCV}, 2018.

\bibitem{wu2018blockdrop}
Zuxuan Wu, Tushar Nagarajan, Abhishek Kumar, Steven Rennie, Larry~S Davis,
  Kristen Grauman, and Rogerio Feris.
\newblock Blockdrop: Dynamic inference paths in residual networks.
\newblock In {\em CVPR}, 2018.

\bibitem{ying2017depth}
Chris Ying and Katerina Fragkiadaki.
\newblock Depth-adaptive computational policies for efficient visual tracking.
\newblock In {\em EMMCVPR}, 2017.

\bibitem{zhang2018shufflenet}
Xiangyu Zhang, Xinyu Zhou, Mengxiao Lin, and Jian Sun.
\newblock Shufflenet: An extremely efficient convolutional neural network for
  mobile devices.
\newblock In {\em CVPR}, 2018.

\bibitem{zoph2018learning}
Barret Zoph, Vijay Vasudevan, Jonathon Shlens, and Quoc~V Le.
\newblock Learning transferable architectures for scalable image recognition.
\newblock In {\em CVPR}, 2018.

\end{thebibliography}
